\documentclass[letterpaper, 10 pt, conference]{ieeeconf}  % Comment this line out if you need a4paper
\usepackage{graphicx}
\usepackage{amsmath}
\usepackage{amssymb}
\usepackage{graphicx}
\usepackage{algorithm}
\usepackage{algpseudocode}
\usepackage{multirow}
\IEEEoverridecommandlockouts                              % This command is only needed if 
\title{\LARGE \bf
RoamFlow: Reinforcement-Aligned One-Step Action MeanFlow Policy for Image-Goal Navigation
}

\author{%
\authorblockN{Zixuan Zhang$^{*}$, Yuqi Chen$^{*}$, Junjie Gao, Siyuan Song, Yongzhou Pan, Beichen Wang, Mir Feroskhan$^{\dagger}$}
\authorblockA{Nanyang Technological University}%
\thanks{$^{*}$Equal contribution. $^{\dagger}$Corresponding author.}%
}

\author{Zixuan Zhang$^{*}$, Yuqi Chen$^{*}$, Junjie Gao, Siyuan Song, Yongzhou Pan, Beichen Wang, and Mir Feroskhan$^{\dagger}$% <-this % stops a space
\thanks{$^{*}$Zixuan Zhang and Yuqi Chen contributed equally to this work.}% <-this % stops a space
\thanks{$^{\dagger}$Mir Feroskhan is the corresponding author.}% <-this % stops a space
\thanks{Zixuan Zhang, Yuqi Chen, Junjie Gao, Siyuan Song, Yongzhou Pan, Beichen Wang, Mir Feroskhan are with Nanyang Technological University, Singapore.
}
}

\begin{document}

\maketitle
\thispagestyle{empty}
\pagestyle{empty}

%%%%%%%%%%%%%%%%%%%%%%%%%%%%%%%%%%%%%%%%%%%%%%%%%%%%%%%%%%%%%%%%%%%%%%%%%%%%%%%%
\begin{abstract}

Image-goal navigation is a key challenge in embodied robotics, where an agent must reach a target specified solely by a goal image. While existing reinforcement learning approaches map perceptual observations directly to actions, they struggle to model long-horizon dependencies, often leading to suboptimal trajectories. 
To address this limitation, we propose \textbf{RoamFlow}, a generative navigation framework that leverages MeanFlow to predict the average velocity field for trajectory synthesis, enabling efficient few-step generation and reducing inference latency. We further adopt a two-stage training strategy that combines expert imitation for stable initialization with reinforcement learning for task-specific policy refinement. 
Extensive experiments in both Habitat simulation and real-world robotic platforms demonstrate that RoamFlow achieves efficient inference while maintaining strong navigation performance under real-time constraints.
\end{abstract}

%%%%%%%%%%%%%%%%%%%%%%%%%%%%%%%%%%%%%%%%%%%%%%%%%%%%%%%%%%%%%%%%%%%%%%%%%%%%%%%%
\section{INTRODUCTION}
Image-goal navigation \cite{imagegoal} is a central problem in embodied robotics, where an agent must reach a target location specified solely by a goal image using onboard sensory observations. Existing reinforcement learning approaches \cite{sign,memory,ppo-cma} typically adopt end-to-end policies that map perceptual observations directly to actions. While such methods demonstrate robustness to perceptual noise and environment uncertainty, their step-wise action prediction paradigm struggles to capture long-horizon dependencies. As a result, these policies often exhibit myopic behaviors and produce suboptimal trajectories in complex environments where multi-step planning and global reasoning are required.

To address this limitation, recent studies reformulate navigation as a trajectory distribution modeling problem within generative frameworks. Instead of predicting single-step actions, generative policies synthesize coherent multi-step action sequences, enabling trajectory-level reasoning and alleviating short-sighted decision-making. Representative approaches include diffusion-based methods such as NoMaD \cite{nomad} and conditional flow matching methods such as FlowNav \cite{flownav}. By modeling a distribution over trajectories, these methods improve long-horizon coherence and allow multimodal trajectory prediction.

\begin{figure}[t]
    \centering
    \includegraphics[width=1\linewidth]{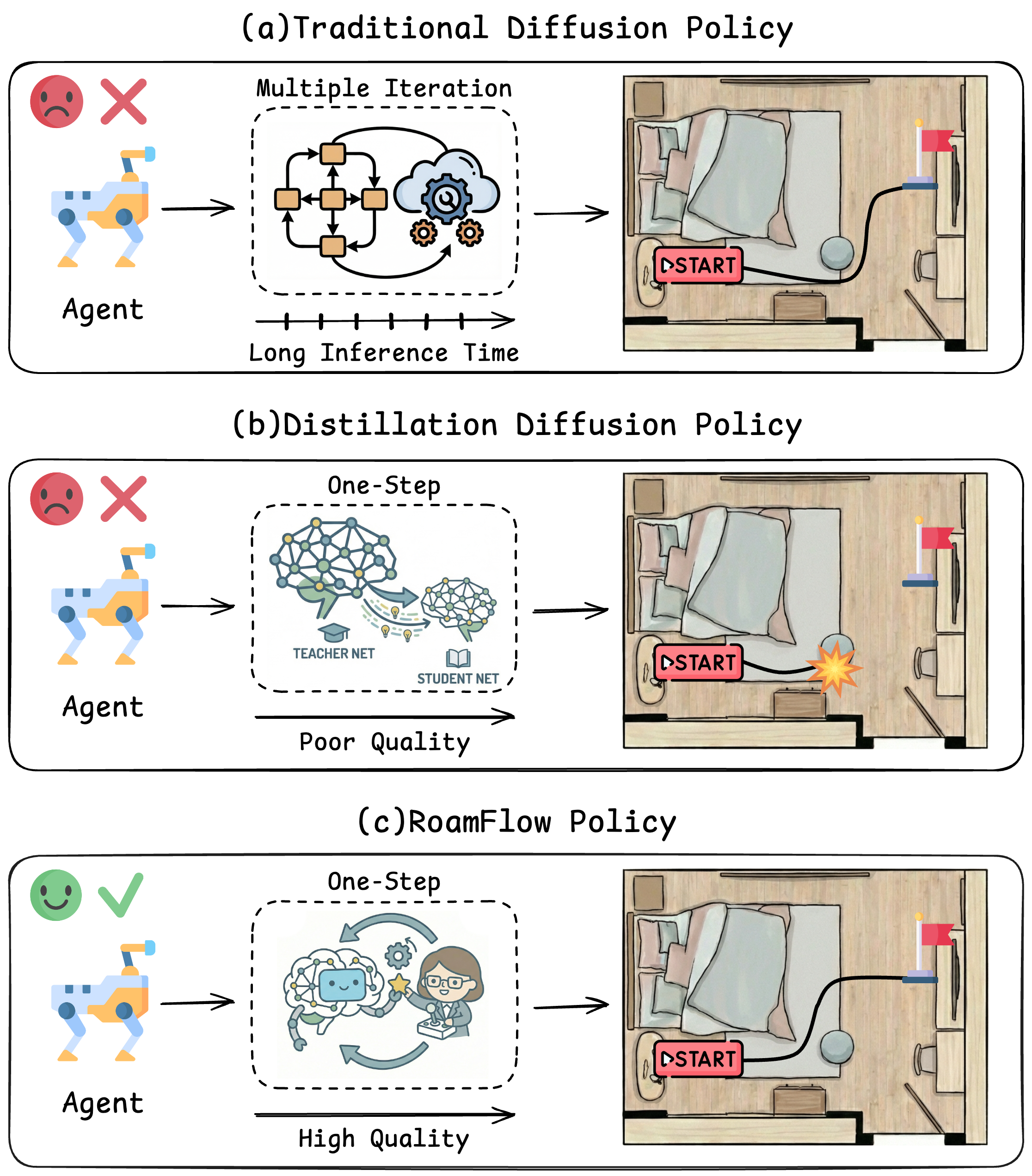}
    \caption{Comparison of different generative policies for image-goal navigation: (a) Traditional Diffusion Policy, which requires multiple iterations, leading to long inference time; (b) Distillation Diffusion Policy, which reduces inference time to a single step but sacrifices trajectory quality; (c) RoamFlow Policy, the proposed method, which utilizes a single-step generation mechanism while maintaining high-quality performance.}
    \label{fig:overview}
\end{figure} 

\begin{figure*}[t]
  \centering
  \includegraphics[width=\textwidth]{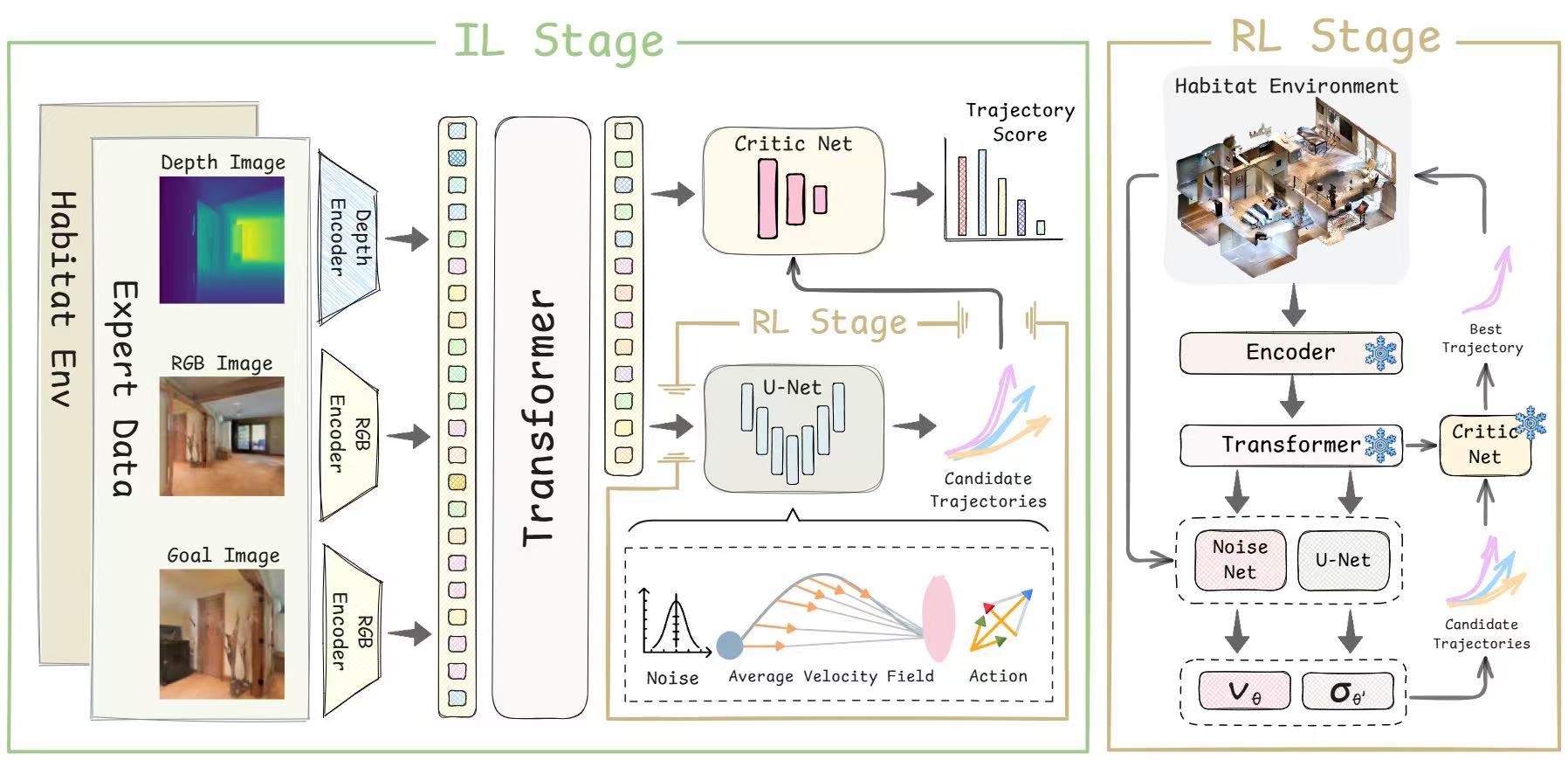}
  \caption{Overview of the two-stage training pipeline. The policy is pretrained via imitation learning (IL) and then fine-tuned with reinforcement learning (RL). The current RGB-D observation and the goal image are encoded and fused into a joint latent representation, which conditions a MeanFlow-based trajectory generator to produce candidate trajectories and a trajectory evaluator head to select safe, goal-directed behaviors.}
  \label{fig:method-overview}
\end{figure*}

However, despite these advantages, generative navigation policies still face two fundamental challenges. 
First, their inference efficiency remains limited by inherently iterative generation processes. 
Diffusion-based policies require multiple denoising steps to generate each trajectory, while flow-based methods still depend on step-wise numerical integration of learned velocity fields. 
Such iterative procedures introduce non-negligible computational latency, hindering real-time deployment on resource-constrained robotic platforms. 
Distillation-based acceleration methods \cite{one,consistency,var} attempt to reduce inference steps by approximating multi-step teacher distributions, but aggressive compression often degrades trajectory quality and diversity.
Second, most generative navigation policies rely primarily on imitation learning. 
Although offline demonstrations provide useful trajectory priors, imitation alone cannot correct suboptimal expert behaviors or adapt to unseen environments during deployment. 
As a result, policies trained purely with imitation may accumulate long-horizon errors and fail to align trajectory generation with task-level objectives such as efficient goal reaching and collision avoidance.

Motivated by these limitations, we propose RoamFlow, a generative navigation framework that combines efficient trajectory generation with reinforcement learning for task-driven refinement. RoamFlow leverages the MeanFlow \cite{mean} model to predict the average velocity field that transports Gaussian noise to the target action sequence. By modeling interval-level displacement instead of instantaneous velocity, the proposed policy significantly reduces the number of required transport steps, leading to more efficient inference while preserving trajectory quality.
To further enhance task performance, RoamFlow adopts a two-stage training strategy. The policy is first initialized via offline imitation learning to capture expert navigation priors. Subsequently, reinforcement learning is applied in the Habitat simulation environment \cite{habitat} to refine the trajectory policy using task-specific rewards, enabling the agent to  better align trajectory generation with navigation objectives.

In this work, our contributions are summarized as follows:
\begin{itemize}
    \item We propose a MeanFlow-based generative navigation policy that predicts the average velocity field, enabling efficient few-step trajectory generation and reducing inference latency.

    \item We introduce a two-stage training framework that combines offline imitation learning for stable initialization with reinforcement learning for task-driven policy refinement.

    \item We validate the proposed framework in both Habitat simulation and real-world robotic experiments, demonstrating efficient inference and strong navigation performance.
\end{itemize}

\section{Related Work}

\subsection{Generative Policies for Navigation}

Recent studies have explored diffusion models~\cite{ddpm} for trajectory generation in navigation tasks, where policies iteratively transform Gaussian noise into structured trajectories, effectively capturing complex, multi-modal action distributions. For example, NoMaD~\cite{nomad} formulates navigation as a diffusion-based policy and generates multi-modal action sequences conditioned on visual observations, improving robustness in previously unseen environments. NaviDiffusor~\cite{navidiffusor} further incorporates gradient guidance derived from environmental constraints and task costs during the diffusion process, thereby enhancing the safety and feasibility of the generated trajectories. NaviBridger~\cite{prior} introduces a diffusion bridge~\cite{ddbm} formulation that conditions the generative process on an informative prior action distribution rather than pure Gaussian noise, which improves generation stability and facilitates long-horizon trajectory synthesis.
To further improve sampling efficiency, recent work has investigated flow-based generative formulations for navigation. In particular, FlowNav~\cite{flownav} adopts Conditional Flow Matching (CFM)~\cite{flow} to learn continuous transport dynamics in trajectory space, enabling more efficient trajectory generation while preserving expressive modeling capacity.
Despite these advances in trajectory modeling, most existing methods still rely on iterative sampling or numerical integration during inference, which introduces considerable computational overhead and poses challenges for real-time deployment on resource-constrained robotic platforms.

\subsection{Fast Sampling for Diffusion Models}

Reducing the inference cost of generative models has become an important research direction, motivating a number of approaches that aim to reduce the number of sampling steps. A common strategy is knowledge distillation~\cite{knowledge,onem}, which compresses a multi-step generative process into a few-step or single-step generator. For example, Progressive Distillation~\cite{progressive} progressively compresses the sampling procedure of diffusion models, enabling significantly faster generation. Distribution Matching Distillation (DMD)~\cite{one} further learns a one-step generator by matching the output distribution of the original diffusion model. In addition, Consistency Models~\cite{consistency1} learn a consistency mapping that enables few-step or even single-step sampling while preserving the learned data distribution. Beyond distillation-based approaches, recent work has explored intrinsically one-step generative formulations. In particular, MeanFlow models~\cite{mean} the average velocity field of the transport dynamics between distributions, allowing samples to be generated with a single forward pass of the network. 

\subsection{Reinforcement Learning for Generative Policies}

Deep reinforcement learning (DRL)\cite{ppo,sac} has recently been explored to further improve generative policies beyond imitation learning. Diffusion policies are typically trained using offline data, which limits their performance when the expert dataset does not fully cover the state space. To address this limitation, several studies incorporate DRL to refine pre-trained generative policies using task-driven reward signals. For example, Diffusion Policy Policy Optimization (DPPO)\cite{dppo} formulates the diffusion denoising process as a Markov decision process and applies policy gradient methods for policy improvement. FDPP~\cite{fdpp} refines diffusion policies by using human preference feedback, learning a reward model to align the policy with subjective tasks while maintaining task performance. NCDPO~\cite{ncdpo} optimizes the policy in the noise-conditioned space, applying DRL with backpropagation through diffusion timesteps, enabling efficient adaptation without modifying the original architecture. In addition, ReinFlow~\cite{reinflow} proposes an online DRL framework for fine-tuning flow-matching policies, allowing reward-aware trajectory optimization for robotic control tasks. 

\section{METHODOLOGY}

In this section, we first provide an overview of the RoamFlow architecture. The proposed framework is trained in two stages. In the first stage, a MeanFlow-based generator and an additional evaluation head for safety-aware trajectory selection are trained via imitation learning. In the second stage, reinforcement learning is used to further refine the navigation policy, improving task-level navigation performance.

\subsection{RoamFlow Architecture}

Our architecture is illustrated in Figure~\ref{fig:method-overview}. The policy maps the current observation to future control actions. Let \( o_t \) denote the policy observation at time \( t \), and \( a_t \) represent a single action. We define a trajectory action sequence as \( A_t = \{a_t, \ldots, a_{t+T-1}\} \). The architecture consists of three main components:

\paragraph{Observation Encoding}

The policy operates on the current RGB image $I_{\text{obs}}^{rgb}$, depth image $I_{\text{obs}}^{depth}$, and goal RGB image $I_{\text{goal}}^{rgb}$. 
We first encode $I_{\text{obs}}^{rgb}$ and $I_{\text{obs}}^{depth}$ with two separate EfficientNet-B0 backbones~\cite{efficentnet} to obtain two corresponding embeddings, which are then concatenated and linearly projected into a joint observation feature. 
In parallel, we concatenate $I_{\text{obs}}^{rgb}$ and $I_{\text{goal}}^{rgb}$ and feed the resulting input into another EfficientNet-B0 encoder to generate a goal embedding. These two embeddings are concatenated and then fed into the Transformer~\cite{attention} block to derive a latent representation \( C_t \).

\paragraph{RoamFlow Policy}
RoamFlow Policy leverages the output token \( C_t \) from the Transformer and incorporates observation conditions through FiLM (Feature-wise Linear Modulation)~\cite{film}. The system takes Gaussian noise as input and uses a UNet~\cite{unet} to predict the mean velocity field that maps the noise distribution to the target distribution. Unlike traditional autoregressive action prediction, trajectory generation is formulated as a single-step process, directly mapping the stochastic noise to a trajectory sequence \( A_t \). This design not only enhances generation efficiency but also ensures that the generated trajectories are effectively conditioned on specific observations and task requirements.

\paragraph{Trajectory Evaluator}
We employ a multi-layer perceptron (MLP) as the trajectory evaluator head to evaluate the feasibility of candidate trajectories. The input to this  trajectory evaluator consists of visual token representations and multiple trajectories generated by the RoamFlow Policy. By leveraging both the shared visual features and trajectory information, the trajectory evaluator assigns a score to each trajectory. During inference, the trajectory with the highest score is selected as the final output. This mechanism enhances robustness in multimodal trajectory prediction by effectively selecting the optimal trajectory from multiple candidates, thereby mitigating potential instability arising from the diversity in trajectory predictions.

Together, these components constitute a cohesive system that efficiently encodes observations, generates trajectories, and evaluates their feasibility, enabling robust image-goal navigation.

\subsection{Offline Policy Learning}\label{sec:offline-policy-learning}
\noindent\textbf{MeanFlow Generator.}
The traditional CFM approach formulates trajectory generation as a continuous transport process that maps a noise prior to expert navigation trajectories.  
Following such paradigm, the trajectory state \( x_t \) evolves according to an ordinary differential equation (ODE):
\begin{equation}
\frac{d}{dt} x_t = v(x_t,t),
\end{equation}
Where \( v(x_t,t) \) denotes the instantaneous velocity field, governing the generation of the trajectory over time. Starting from a Gaussian noise sample \( x_0 \sim \mathcal{N}(0, I) \), the trajectories are obtained by integrating this velocity field, transforming the noise distribution into expert navigation trajectories.

However, instantaneous velocity formulations require numerical integration of the velocity field, typically involving multiple transport steps during inference. This increases computational cost and limits real-time deployment.
To address this issue, we adopt MeanFlow, which predicts the average velocity field instead of the instantaneous velocity. By modeling interval-level displacement, MeanFlow enables efficient trajectory generation with significantly fewer transport steps while preserving trajectory quality.

MeanFlow predicts interval-level displacement rather than instantaneous velocity. 
Specifically, the average velocity between two time steps $r$ and $t$ is defined as
\begin{equation}
u(x_t,r,t)=
\frac{1}{t-r}\int_{r}^{t} v(x_\tau,\tau)\, d\tau,
\end{equation}
which represents the normalized displacement induced by the transport process. 
By construction, the average velocity is consistent with the instantaneous velocity in the limit $r\rightarrow t$, i.e.,
\begin{equation}
\lim_{r\to t} u(x_t,r,t)=v(x_t,t),
\end{equation}
establishing a principled connection between displacement-based and velocity-based trajectory modeling.

In practice, we train a neural network $u_\theta$ to predict this average velocity field conditioned on observations, enabling trajectory synthesis with a small number of transport steps.
This identity provides a natural training target for learning displacement-aware trajectory dynamics.We therefore optimize $u_\theta$ using the objective
\begin{equation}
\mathcal{L}_{\mathrm{MF}}
=
\mathbb{E}_{x_t,r,t}
\left[
\left\|
u_\theta(x_t,r,t)
-
u(x_t,r,t)
\right\|^2
\right],
\end{equation}
which encourages the policy to capture interval-level motion structure.

By modeling $u$ rather than $v$, the policy enables trajectory synthesis with substantially fewer transport steps, improving inference speed without sacrificing trajectory fidelity, which is particularly important for real-time navigation.

\noindent\textbf{Trajectory Evaluator.}
Given observation $o$ and a candidate trajectory $\tau=\{\tau_k,a_k\}_{k=0}^{T}$, 
the trajectory evaluator predicts a scalar score $s_{\phi}(o,\tau)$ used to rank trajectory hypotheses. 
Supervision is constructed from geometric clearance signals obtained from a distance field. 
Let $d_k=d_{\text{obs}}(\tau_k)$ denote the distance to the nearest obstacle. 
We define a trajectory-level target
\begin{equation}
\tilde{s}(\tau)
=
s_{\text{safe}}(\tau)
-
\lambda\, s_{\text{smooth}}(\tau),
\end{equation}
with
\begin{equation}
s_{\text{safe}}(\tau)
=
-
\sum_{k=0}^{T}
\psi\!\left(d_{\text{safe}}-d_k\right)
+
\alpha
\sum_{k=0}^{T-1}
\left(d_{k+1}-d_k\right),
\end{equation}
\begin{equation}
s_{\text{smooth}}(\tau)
=
\sum_{k=1}^{T-1}
\lVert a_{k+1}-2a_k+a_{k-1}\rVert^2,
\end{equation}
where $\psi(\cdot)$ is a smooth barrier function increasing when clearance falls below a safety threshold $d_{\text{safe}}$. 
The evaluator is trained via regression
\begin{equation}
\mathcal{L}_{\text{eval}}
=
\mathbb{E}_{(o,\tau)}
\big[
\left(s_{\phi}(o,\tau)-\tilde{s}(\tau)\right)^2
\big],
\end{equation}
enabling reliable ranking of candidate trajectories at inference time.

At inference time, the generator produces multiple trajectory proposals and the evaluator selects the trajectory with the highest score
\begin{equation}
\tau^{*}=\arg\max_{\tau_i}s_{\phi}(o,\tau_i),
\end{equation}
providing a safety-aware selection layer that improves robustness under multimodal predictions. 
The evaluator is trained offline and used solely for trajectory selection during deployment.

\begin{algorithm}[t]
\small
\caption{ROAM: Two-Stage Trajectory Policy Optimization}
\label{alg:roam}
\begin{algorithmic}

\State \textbf{Input} expert dataset $\mathcal{D}=\{(o_t,A_t^\star)\}$; environment $\mathcal{E}$; transport steps $K$; discount $\gamma$; batch size $B$; clipping range $\epsilon \in (0, 1)$.

\State \textbf{Initialize} trajectory policy composed of backbone $f_{\psi}(o)\!\to\!Z$, MeanFlow generator $u_{\theta}$, and trajectory evaluator $s_{\phi}(o,\tau\mid Z)$.

\State \textbf{Initialize} MeanFlow generator parameters $\theta$, stochastic noise head $\theta'$, and evaluator parameters $\phi$.

\vspace{0.3em}
\State \textbf{Stage I: Generator and Evaluator Training}
\While{not converged}
    \State Sample $(o,A^\star)\sim\mathcal{D}$ and compute $Z=f_{\psi}(o)$.
    \State Sample $X^0\sim p_{\mathrm{prior}}$, sample $(r,t)$, construct $X^t$.
    \State Update $\theta$ by minimizing $\|u_\theta(X^t,r,t|Z)-u(X^t,r,t)\|^2$.
    \State Generate trajectory proposals $\{\tau_i\}$ and compute targets $\tilde{s}(\tau_i)$.
    \State Update $\phi$ by regression on $s_\phi(o,\tau_i|Z)$.
\EndWhile

\vspace{0.3em}
\State \textbf{Stage II: Online Policy Refinement}
\State Freeze Transformer parameters $\psi$ and evaluator $\phi$.
\While{not converged}
    \State $\bar{\theta}_{old} \leftarrow \bar{\theta}$
    \State Reset environment and observe $o$.
    \While{episode not terminated}
        \State $Z \leftarrow f_{\psi}(o)$; \quad $X^0 \sim \mathcal{N}(0,I)$
        \State Generate trajectory $\tau$ via stochastic MeanFlow transport
        \State Execute $\tau$, observe $(r,d,o)$, and store transition in buffer
    \EndWhile
    \State Sample mini-batch; evaluate $\log \pi_{\bar{\theta}}(\tau|o)$
    \State Estimate advantages and update $\bar{\theta}$ with clipped PPO
\EndWhile

\end{algorithmic}
\end{algorithm}

\subsection{Online Policy Refinement}

While the offline stage learns a trajectory prior from expert demonstrations, imitation alone cannot correct suboptimal behaviors or long-horizon decision errors in real navigation. Therefore, we further incorporate online reinforcement learning to refine the navigation policy. we formulate trajectory generation as a finite-horizon stochastic transport process over latent trajectory states. 
Let $\{X_t^k\}_{k=0}^{K}$ denote the intermediate trajectory states, where $X_t^K=\tau_t$ is the final trajectory executed in the environment. 
This formulation treats trajectory synthesis as a discrete-time Markov process whose transitions define a tractable trajectory distribution.

However, the deterministic MeanFlow formulation in Eq.~(2) specifies only a transport mapping and does not provide an explicit stochastic transition model, preventing likelihood-based policy optimization. 
To obtain a stochastic process, we introduce learnable noise into the transport dynamics, where the transition variance is predicted by a learnable noise network $g_{\theta'}(\cdot)$, resulting in Gaussian transitions conditioned on the current latent state.
\begin{equation}
\begin{aligned}
X_t^{k+1}
&\sim
\mathcal N\!\Big(
X_t^k + U_\theta(X_t^k, Z_t, t_k)\Delta t_k,\;
\Sigma_{\theta'}(X_t^k, Z_t, t_k)
\Big),
\end{aligned}
\end{equation}
with initialization $X_t^0\sim\mathcal N(0,I)$.
This converts deterministic transport into a stochastic Markov chain over trajectory latents.

Because each transition is reparameterized, the trajectory distribution is factorized as a product of transition densities,
\begin{equation}
\begin{aligned}
\log \pi_{\bar\theta}(\tau_t \mid o_t)
&=
\log \mathcal N(X_t^0; 0, I) \\
&\quad +
\sum_{k=0}^{K-1}
\log \mathcal N\!\Big(
X_t^{k+1};
\mu_{\bar\theta}^k,\,
\Sigma_{\theta'}^k
\Big),
\end{aligned}
\end{equation}
where $\mu_{\bar\theta}^k
=
X_t^k + U_\theta(X_t^k, Z_t, t_k)\Delta t_k$
and $\bar\theta=\{\theta,\theta'\}$.

At inference time, the stochastic transport chain collapses into a deterministic policy mapping, which directly maps an initial latent sample to a trajectory:
\begin{equation}
\tau_t
=
f_{\theta}(X_t^{0}, Z_t),
\
f_{\theta}(X,Z)
=
X + U_{\theta}(X,Z).
\end{equation}

At each environment step, the reward is defined as
\begin{equation}
r_t
=
r^{\text{success}}_t
+
r^{\text{slack}}_t
+
r^{\text{progress}}_t
+
r^{\text{collision}}_t,
\end{equation}
where $r^{\text{success}}_t$ is a positive reward assigned when the agent reaches the goal at step $t$,
$r^{\text{slack}}_t$ is a constant per-step penalty to encourage faster task completion,
$r^{\text{progress}}_t = \Delta d_t$ measures the change in distance to the goal between two consecutive steps, 
$r^{\text{collision}}_t$ is a penalty term applied when a collision occurs at step $t$.
The episode return is
\begin{equation}
G_t=\sum_{j=0}^{T-1}\gamma^j r_{t+j},
\end{equation}
and the trajectory-level advantage $A_t$ is estimated from the collected rollouts.

The trajectory generation policy is optimized using the clipped PPO objective
\begin{equation}
\mathcal{L}_{\mathrm{PPO}}
=
\mathbb{E}_t\!\left[
\min\!\left(
r_t A_t,\;
\mathrm{clip}(r_t,1-\epsilon,1+\epsilon)A_t
\right)
\right],
\end{equation}
where the importance ratio is
$
r_t
=
\frac{
\pi_{\bar{\theta}}(\tau_t \mid o_t)
}{
\pi_{\bar{\theta}_{\mathrm{old}}}(\tau_t \mid o_t)
}.
$

%%%%%%%%%%%%%%%%%%%%%%%%%%%%%%%%%%%%%%%%%%%%%%%%%%%%%%%%%%%%%%%%%%%%%%%%%%%%%%%%%%%%%%%%%%%%%%%%%%%%%%%%%%%%%%%%%%%%%%%%%%%%%%%%%%%%%%%%%%%%%%%%%%%%%%%%%%%%%%%
\section{Experiments}

\subsection{Simulation Setup}

We formulate the image-goal navigation task as a goal-conditioned policy learning problem. 
Without global localization, the agent navigates on a pre-built topological graph. 
In each episode, given a distant goal image $I_g$, the agent performs self-localization on the map to select the next sub-goal image. 
The RoamFlow policy then predicts trajectory waypoints, which are tracked by a PD controller to produce velocity commands for goal-directed navigation.

For Habitat, we use an RGB-D camera with a $90^\circ$ horizontal field of view for perception. 
The action space consists of continuous linear and angular velocities $(v_t, \omega_t)$. 
An episode is considered successful when the agent issues the \textit{Stop} action and its final pose is within $d_s = 1.0$ m in position and within $\alpha_s = 30^\circ$ in heading relative to the target observation pose.

We evaluate navigation using Success Rate (SR), Success weighted by Path Length (SPL), Collision Rate (CR), and Inference Time (ms). 
A collision is recorded whenever the simulator reports contact, and CR is computed as the proportion of episodes containing at least one collision.

%%%%%%%%%%%%%%%%%%%%%%%%%%%%%%%%%%%%%%%%%%%%%%%%%%%%%%%%%%%%%%%%%%%%%%%%%%%
% B Model training

\subsection{Training Details}

\subsubsection{Datasets}
We utilize a diverse collection of indoor environments for training and evaluation, including the Gibson \cite{gibson} and Matterport3D (MP3D) \cite{mp3d} simulation datasets. 
During data collection, we run Hybrid A* on the scene NavMesh to generate smooth and collision-free expert trajectories. 
The planned trajectories are temporally aligned with the recorded RGB-D observations, forming an expert trajectory dataset. 
The RGB image at the terminal step of each trajectory is used as the episode goal image.

We further incorporate episodes from real-world datasets GoStanford \cite{gostanford} and Scand \cite{scand} to diversify the training data and reduce the simulation-to-reality gap. 
As the GoStanford dataset lacks native depth information, we employ Depth Anything V2 \cite{depth_any_thing} to generate high-fidelity depth maps.

\subsubsection{Imitation Learning}
Thee model is first pre-trained with imitation learning using the AdamW optimizer, a learning rate of $1\times 10^{-5}$
, a batch size of 128, and 25 training epochs. The training is conducted on an NVIDIA RTX 6000 Ada GPU for 20 hours.

\subsubsection{Reinforcement Learning}
We then fine-tune the pretrained policy in Habitat. Exploration noise is injected with standard deviations ranging from 0.08 to 0.14. We adopt PPO with a discount factor of 0.99, a GAE parameter of 0.95, and a batch size of 1000. The reward function consists of a success reward of 5, a per-step slack penalty of $-0.01$, and a collision penalty of $-0.1$. All RL experiments are conducted on the same GPU for 30 hours.
%%%%%%%%%%%%%%%%%%%%%%%%%%%
%reward图

\begin{figure}[!t]
  \centering
  \includegraphics[width=\columnwidth]{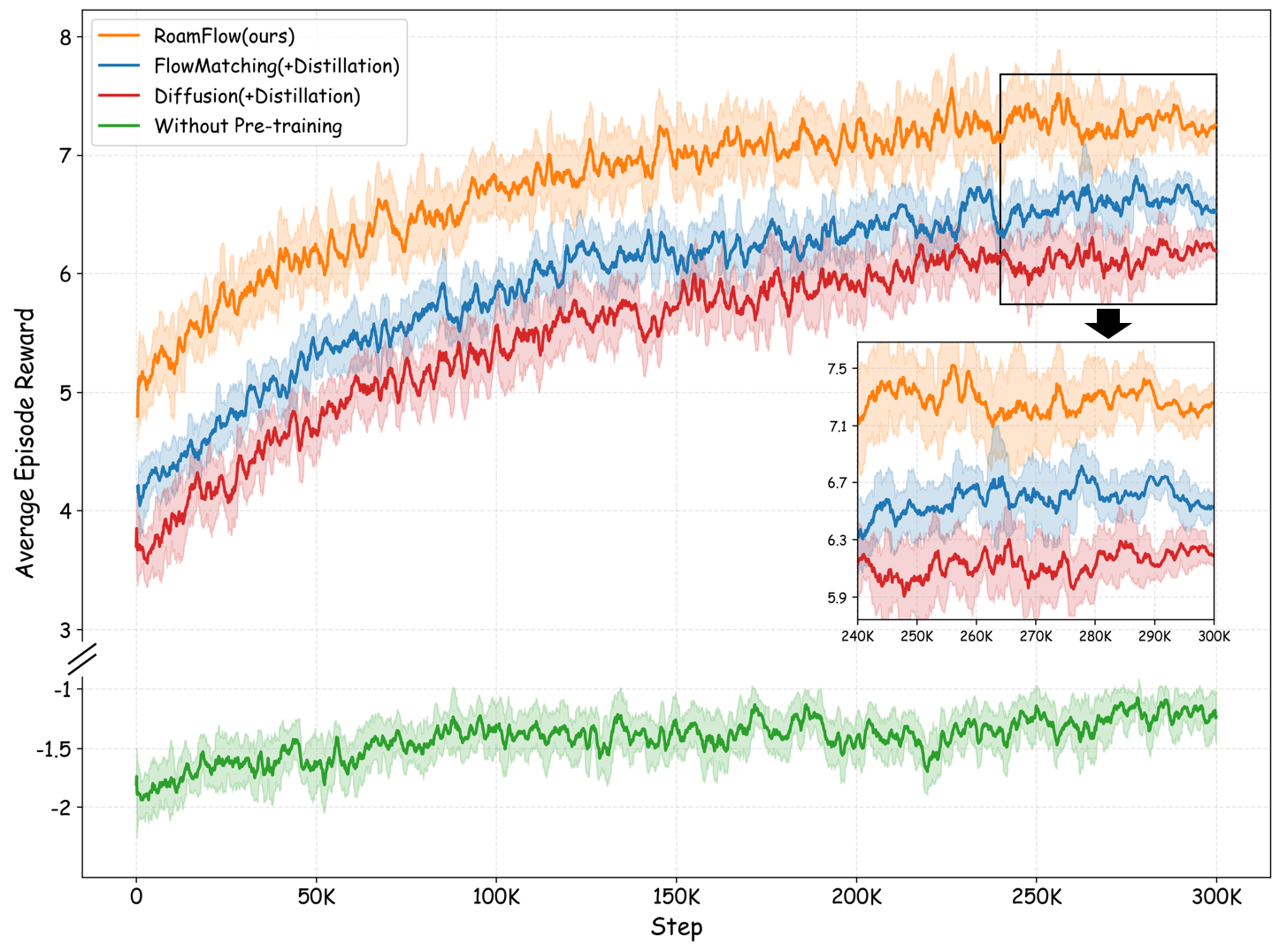}
  \caption{Reward curve during training.The steady rise during the RL stage demonstrates effective policy alignment with task objectives.}
  \label{fig:reward_success}
\end{figure}

%%%%%%%%%%%%%%%%%%%%%%%%%%%%%%%%%%%%%%%%%%%%%%%%%%%%%%%%%%%
%大表
%%%%%%%%%%%%%%%%%%%%%%%%%%%%%%%%%%%%%%%%%%%%%%%%%%%%%%%%%%%
\begin{table*}[!t]
\caption{Comparison between RoamFlow and baselines on Gibson and MP3D. All methods are trained on the Gibson training split. Results on the unseen Gibson test scenes reflect in-domain performance, whereas MP3D evaluates cross-domain generalization without finetune.}
\label{tab:baseline_comparison}
\centering
\small
\setlength{\tabcolsep}{3.5pt}
\resizebox{\textwidth}{!}{%
\begin{tabular}{lcccccc|cccc}
\hline
\multirow{2}{*}{Method} & \multirow{2}{*}{\shortstack{Depth\\Input}} & \multirow{2}{*}{\shortstack{Trajectory\\Selection}} & \multicolumn{4}{c|}{Gibson} & \multicolumn{4}{c}{MP3D} \\
\cline{4-11}
 &  &  & SR(\%)$\uparrow$ & SPL(\%)$\uparrow$ & CR(\%)$\downarrow$ & Time(ms)$\downarrow$ & SR(\%)$\uparrow$ & SPL(\%)$\uparrow$ & CR(\%)$\downarrow$ & Time(ms)$\downarrow$ \\
\hline
NoMaD \cite{nomad} & No & Random & 43.1 & 29.4 & 17.3 & 49.1 & 34.8 & 19.7 & 25.1 & 49.4 \\
NoMaD \cite{nomad} + Distillation  & No & Random & 38.4 & 25.3 & 21.3 & 28.1 & 30.4 & 13.5 & 27.8 & 29.1 \\
FlowNav \cite{flownav} & Yes & Random & 50.7 & 34.1 & 16.1 & 47.7 & 35.1 & 24.1 & 20.3 & 46.1 \\
FlowNav \cite{flownav} + Distillation  & Yes & Random & 47.5 & 29.7 & 20.1 & 21.2 & 30.2 & 19.2 & 24.3 & 20.2 \\
NaviDiffusor \cite{navidiffusor} & Yes & Cost-Guided & 51.1 & 45.4 & 11.8 & 89.5 & 46.1 & 34.6 & 13.3 & 90.1 \\
NaviBridger \cite{prior} & No & N/A & 47.4 & 37.6 & 15.8 & 44.3 & 40.8 & 23.3 & 20.5 & 43.3\\
NavDp \cite{navdp} & Yes & Critic-Guided & 59.3 & 47.5 & 11.6 & 61.1 & 48.1 & 34.1 & 17.5 & 62.5 \\
\hline
\textbf{RoamFlow (Ours)} & \textbf{Yes} & \textbf{Critic-Guided} & \textbf{68.7} & \textbf{61.9} & \textbf{10.9} & \textbf{19.6} & \textbf{56.1} & \textbf{47.1} & \textbf{12.2} & \textbf{19.1} \\
\hline
\end{tabular}%
}
\end{table*}

%大表
%%%%%%%%%%%%%%%%%%%%%

%%%%%%%%%%%%%%%%%%%%%%%%%%%%%%%%%%%%%%%%%%%%%%%%%%%%%%%%%%%%%%%%%%%%%%%%
% C、baseline

\subsection{Baseline}
We compare RoamFlow with the following baselines in the Habitat simulator: 
\subsubsection{Generative Navigation Baselines}
Diffusion-based navigation policies (NoMaD \cite{nomad}, NaviDiffusor \cite{navidiffusor}, and NavDP \cite{navdp}) typically rely on iterative diffusion sampling to generate action or trajectory sequences, which can model multimodal behaviors but often incurs substantial inference overhead.
FlowNav \cite{flownav} replaces diffusion sampling with Conditional Flow Matching, achieving comparable performance with fewer inference steps.
NaviBridger \cite{prior} starts from informative prior actions and leverages a diffusion bridge to gradually transport the source distribution toward the target action distribution, improving the efficiency-performance trade-off by reducing unnecessary sampling steps.

\subsubsection{Distilled Generative Policies}

For diffusion distillation, we adopt a progressive distillation ~\cite{progressive}, compressing a teacher diffusion policy with multi-step denoising into a student that performs few-step update. 
We additionally implement a baseline that distills a pretrained flow-matching policy into a one-step generator using score distillation~\cite{distfm}.

%%%%%%%%%%%%%%%%%%%%%%%%%%%%%%%%%%%%%%%%%%%%%%%%%%%%%%%%%%%%%%%%%%%%%%%%
% D、result

\subsection{Simulation Result}

\begin{figure}[!t]
  \centering
  \includegraphics[width=\columnwidth]{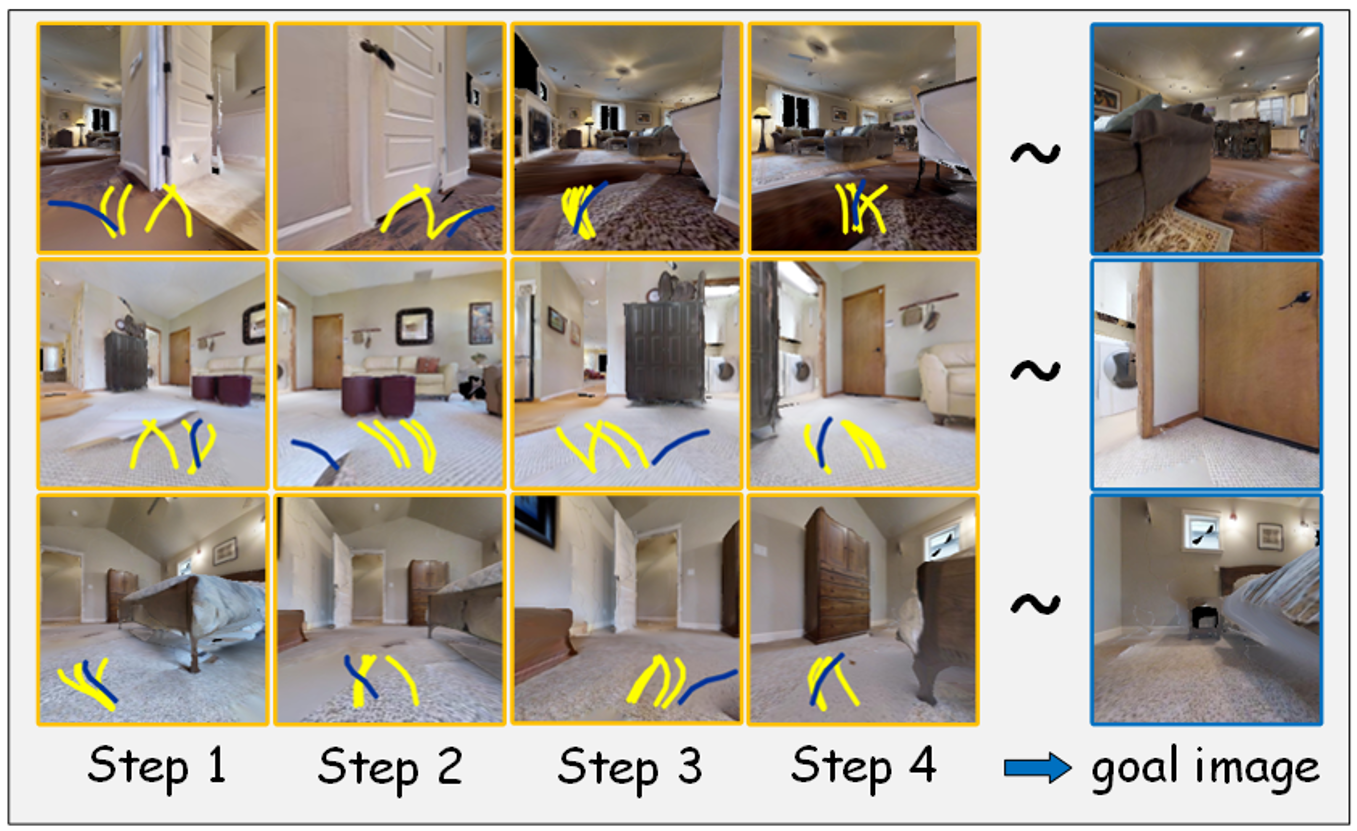}
  \caption{Navigation examples in simulation. At each step, our policy generates several candidate trajectories and selects one to execute (highlighted in blue). The selected trajectory is consistently safe and goal-directed.}
  \label{fig:simulation_test}
\end{figure}

Fig.~\ref{fig:simulation_test} illustrates representative simulation rollouts. At each step, our policy proposes multiple candidate trajectories and executes the safe, goal-directed option, resulting in smooth and collision-averse navigation in cluttered scenes.

%%%%%%%%%%%%%%%%%%%%%%%%%%%
%%仿真测试图

\subsubsection{Baseline Comparison}

To ensure a fair comparison, we evaluate the proposed method and baselines under a unified experimental protocol. All models are trained on the same dataset split and tested in identical environments. During evaluation, identical goal image and pre-built topological graph are provided to all methods. Notably, the originally mapless baselines (NavDP \cite{navdp} and NaviBridger \cite{prior}), are adapted to function as local planners within topological framework. For methods that do not natively require depth, we evaluate them without providing depth inputs and explicitly mark these settings in the results.

From Table~\ref{tab:baseline_comparison}, we can see that RoamFlow achieves the highest performance on the Gibson validation split. Compared with the strongest generative baseline, NavDP, it increases SR by 9.4 and SPL by 14.4, while reducing CR from 11.6 to 10.9 and reducing inference time from 61.1ms to 19.6ms. These results demonstrate that RoamFlow improves navigation quality, efficiency and safety.

 We also demonstrate the reward progression during the RL stage in Fig.~\ref{fig:reward_success}. The average episode reward increases steadily over training, indicating that RL fine-tuning consistently improves policy performance. Notably, RoamFlow (orange) achieves the highest reward throughout training with faster and more stable gains, suggesting quicker convergence and superior final policy quality. Among the distilled baselines, FlowMatching distillation (blue) consistently outperforms diffusion distillation (red), while training without pre-training (green) remains at negative reward values.

\begin{figure*}[!t]
  \centering
  \includegraphics[width=\textwidth]{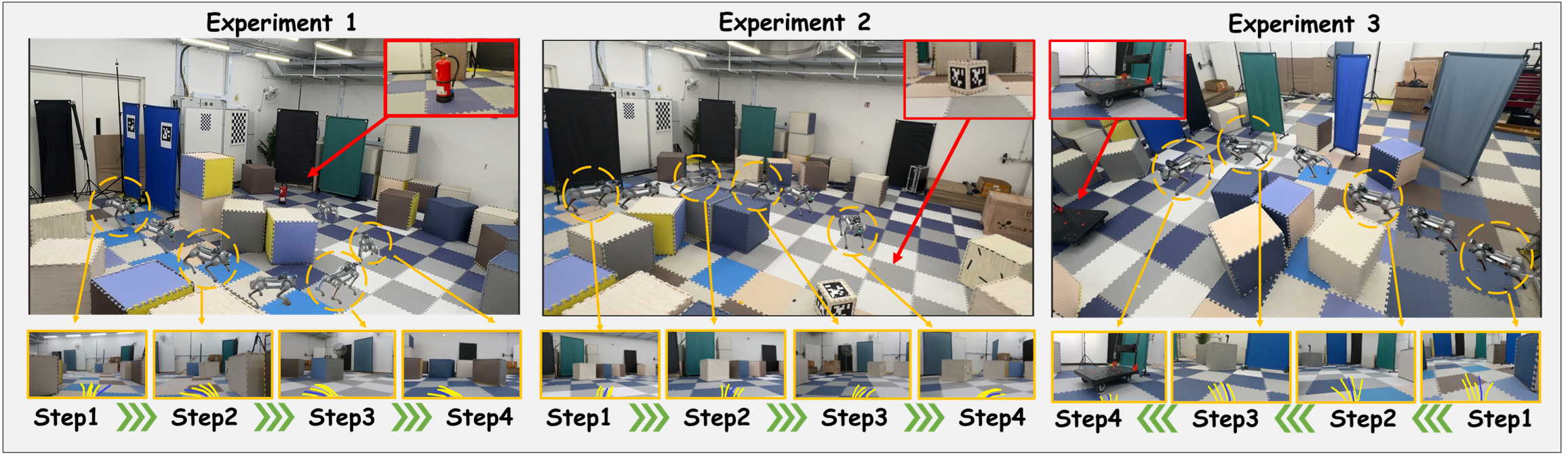}
  \caption{Deployment on real robots. This figure shows RoamFlow rollouts in three scenarios. The red boxed image indicates the goal image. The orange boxed image visualizes the trajectory prediction process, where the blue trajectory denotes the selected and executed trajectory.}
  \label{fig:deployment2}
\end{figure*}

\subsubsection{Ablation Study}
Our ablation study evaluates the impact of three factors: imitation learning (IL) pretraining, reinforcement learning (RL) fine-tuning, and the use of a trajectory evaluator for selecting candidate trajectories.

\paragraph{Effect of IL Stage}
As shown in Table~\ref{tab:ablation_gibson_mp3d}, IL pre-training shows clear gains over training from scratch. Without IL, the agent must learn basic collision-avoidance behaviors via trial-and-error, leading to slower and less stable learning. IL bootstraps these skills and improves exploration efficiency.

\paragraph{Effect of RL Stage}
As shown in Table~\ref{tab:ablation_gibson_mp3d}, starting from the same IL initialization, RL fine-tuning leads to consistent improvements across both datasets, achieving higher SR and SPL while significantly reducing CR compared to the IL-only variant. These results indicate that RL effectively complements imitation by refining the policy through task-oriented rewards, which enhances robustness and safety via direct environment interaction in cluttered indoor scenes. Consequently, the RL stage serves as an objective-alignment step that bridges the gap between the IL prior and practical deployment requirements, further optimizing the policy toward target evaluation metrics.

\paragraph{Effect of Trajectory Evaluator}
As summarized in Table~\ref{tab:ablation_gibson_mp3d}, introducing the evaluator does not change the underlying policy, but re-ranks a set of candidate rollouts and executes the one with lower estimated risk. Compared to random selection, this simple selection mechanism consistently reduces the CR, indicating improved safety in visually cluttered indoor scenes. Meanwhile, by avoiding collision-induced failures and getting stuck less often, the agent achieves improvement in SR and SPL. Overall, the Trajectory Evaluator acts as a lightweight safety filter that improves trajectory reliability with minimal additional overhead, making the system more suitable for deployment.

%%%%%%%%%%%%%%%%%%%%%%%%%%%
%%消融表

\begin{table}[!htbp]
\caption{Ablation of IL, RL, and Trajectory Evaluator on Gibson and MP3D.}
\label{tab:ablation_gibson_mp3d}
\centering
\footnotesize
\setlength{\tabcolsep}{3pt}
\renewcommand{\arraystretch}{0.95}
\begin{tabular}{lcccccc}
\hline
Method & \multicolumn{3}{c}{Gibson} & \multicolumn{3}{c}{MP3D} \\
\cline{2-7}
 & SR$\uparrow$ & SPL$\uparrow$ & CR$\downarrow$ & SR$\uparrow$ & SPL$\uparrow$ & CR$\downarrow$ \\
\hline
\emph{Effect of IL Stage} \\
w/o IL Stage & 13.2 & 11.2 & 61.3 & 11.2 & 9.8 & 68.7 \\
\emph{Effect of RL Stage} \\
w/o RL Stage & 51.3 & 47.5 & 25.9 & 46.8 & 43.5 & 30.6 \\

\emph{Effect of Trajectory Evaluator} \\
Random Selection & 60.7 & 50.2 & 17.5 & 53.2 & 40.6 & 24.4 \\

\hline
\textbf{Full RoamFlow (Ours)} & \textbf{68.7} & \textbf{61.9} & \textbf{10.9} & \textbf{56.1} & \textbf{47.1} & \textbf{12.2} \\
\hline
\end{tabular}
\end{table}

%%%%%%%%%%%%%%%%%%%%%%%%%%%%%%%%%%%%%%%%%%%%%%%%%%%%%%%%%%%%%%%%%%%%%%%%
% E、Physical Experiments 真机实验

\subsection{Physical Experiments}

For real-robot experiments, our navigation policy is deployed on a Unitree Go2 quadruped equipped with an NVIDIA Jetson Orin NX (16GB). We use ROS1 Noetic as the middleware for sensor data streaming and command communication. An Intel RealSense D435i provides synchronized RGB-D observation inputs.

We evaluate multiple navigation methods across three distinct scenarios under the image-goal navigation task. Each method is tested for a total of 20 runs distributed across the three scenarios to ensure statistical reliability. We report Success Rate (SR), the average number of collisions per run (C/run), and the mean inference time as the primary performance metrics.

Table~\ref{tab:physical_experiments} reports success rate, collisions per run, and inference time. RoamFlow attains a higher SR with fewer collisions and lower inference time.

%%%%%%%%%%%%%%%%%%%%%%%%%%%%
%真机实验表
\begin{table}[!htbp]
  \caption{performance in physical experiments (success rate, collisions and inference time(ms)).}
  \label{tab:physical_experiments}
  \centering
  \small
  \setlength{\tabcolsep}{4pt}
  \begin{tabular}{l|ccc}
    \hline
    Method & SR$\uparrow$ & C/run$\downarrow$ & Time$\downarrow$ \\
    \hline

    NoMaD(with Depth) & 0.75 & 0.50 & 91.1 \\
    NoMaD + Distillation(with Depth)& 0.70 & 0.70 & 67.2 \\
    FlowNav & 0.90 & 0.25 & 89.5\\
    FlowNav + Distillation& 0.80 & 0.35 & 38.6 \\
    \hline
    \textbf{RoamFlow (ours)} & \textbf{1.00} & \textbf{0.10} & \textbf{37.2} \\
    \hline

    \hline

\end{tabular}
\end{table}

Fig.~\ref{fig:deployment2} illustrates three successful navigation trajectories of our policy during execution. As shown, the robot consistently avoids unsafe regions while progressing toward the goal.

During the real-world deployment, the whole navigation and control loop is executed onboard at a frequency of 10 Hz. The inference latency consistently remains below 100 ms. Despite minor jitter in computation timing, the system maintains a stable inference rate, satisfying the real-time constraints required for autonomous navigation.

\setlength{\tabcolsep}{3pt}
\renewcommand{\arraystretch}{0.95}

%%%%%%%%%%%%%%%%%%%%%%%%%%%%%%%%%%%%%%%%%%%%%%%%%%%%%%%%%%%%%%%%%%%%%%%%%%%

\section{CONCLUSIONS}
We presented RoamFlow, a MeanFlow-based generative navigation framework for image-goal navigation that enables efficient one-step trajectory generation under real-time constraints. RoamFlow predicts interval-level displacement via an average velocity field, reducing the number of required transport steps during inference. To improve task-level performance, we adopt a two-stage training strategy that combines imitation learning for stable initialization with reinforcement learning for reward-driven policy refinement. In addition, a lightweight trajectory evaluator re-ranks candidate trajectories to select safe and smooth behaviors at inference time. Extensive experiments in Habitat and real-world environments demonstrate that RoamFlow achieves strong navigation performance with low inference latency, offering a practical solution for deploying generative navigation policies on resource-constrained onboard systems.

\addtolength{\textheight}{0cm}   % This command serves to balance the column lengths

%%%%%%%%%%%%%%%%%%%%%%%%%%%%%%%%%%%%%%%%%%%%%%%%%%%%%%%%%%%%%%%%%%%%%%%%%%%%%%%%

%%%%%%%%%%%%%%%%%%%%%%%%%%%%%%%%%%%%%%%%%%%%%%%%%%%%%%%%%%%%%%%%%%%%%%%%%%%%%%%%



% \section*{ACKNOWLEDGMENT}

% The preferred spelling of the word ÒacknowledgmentÓ in America is without an ÒeÓ after the ÒgÓ. Avoid the stilted expression, ÒOne of us (R. B. G.) thanks . . .Ó  Instead, try ÒR. B. G. thanksÓ. Put sponsor acknowledgments in the unnumbered footnote on the first page.

%%%%%%%%%%%%%%%%%%%%%%%%%%%%%%%%%%%%%%%%%%%%%%%%%%%%%%%%%%%%%%%%%%%%%%%%%%%%%%%%

% References are important to the reader; therefore, each citation must be complete and correct. If at all possible, references should be commonly available publications.

\bibliographystyle{IEEEtran}
\bibliography{reference}

@inproceedings{imagegoal,
  title={Target-driven visual navigation in indoor scenes using deep reinforcement learning},
  author={Zhu, Yuke and Mottaghi, Roozbeh and Kolve, Eric and Lim, Joseph J and Gupta, Abhinav and Fei-Fei, Li and Farhadi, Ali},
  booktitle={2017 IEEE international conference on robotics and automation (ICRA)},
  pages={3357--3364},
  year={2017},
  organization={IEEE}
}

@article{sign,
  title={SIGN: Safety-Aware Image-Goal Navigation for Autonomous Drones via Reinforcement Learning},
  author={Yan, Zichen and Huang, Rui and He, Lei and Guo, Shao and Zhao, Lin},
  journal={IEEE Robotics and Automation Letters},
  volume={11},
  number={2},
  pages={1962--1969},
  year={2025},
  publisher={IEEE}
}

@inproceedings{memory,
  title={Memory-augmented reinforcement learning for image-goal navigation},
  author={Mezghan, Lina and Sukhbaatar, Sainbayar and Lavril, Thibaut and Maksymets, Oleksandr and Batra, Dhruv and Bojanowski, Piotr and Alahari, Karteek},
  booktitle={2022 IEEE/RSJ International Conference on Intelligent Robots and Systems (IROS)},
  pages={3316--3323},
  year={2022},
  organization={IEEE}
}

@inproceedings{ppo-cma,
  title={An Improved Reinforcement Learning-Based UAV Obstacle Avoidance Framework Using PPO-CMA},
  author={Chen, Yuqi and Gao, Junjie and Deng, Yaosheng and Feroskhan, Mir},
  booktitle={2025 IEEE International Conference on Systems, Man, and Cybernetics (SMC)},
  pages={5845--5850},
  year={2025},
  organization={IEEE}
}

@article{nomad,
  title={Nomad: Goal masked diffusion policies for navigation and exploration},
  author={Sridhar, Ajay and Shah, Dhruv and Glossop, Catherine and Levine, Sergey},
  booktitle={2024 IEEE International Conference on Robotics and Automation (ICRA)},
  pages={63--70},
  year={2024},
  organization={IEEE}
}

@article{flownav,
  title={Flownav: Combining flow matching and depth priors for efficient navigation},
  author={Gode, Samiran and Nayak, Abhijeet and Oliveira, D{\'e}bora NP and Krawez, Michael and Schmid, Cordelia and Burgard, Wolfram},
  journal={arXiv preprint arXiv:2411.09524},
  year={2024}
}

@article{one,
  title={One-step diffusion policy: Fast visuomotor policies via diffusion distillation},
  author={Wang, Zhendong and Li, Zhaoshuo and Mandlekar, Ajay and Xu, Zhenjia and Fan, Jiaojiao and Narang, Yashraj and Fan, Linxi and Zhu, Yuke and Balaji, Yogesh and Zhou, Mingyuan and others},
  journal={arXiv preprint arXiv:2410.21257},
  year={2024}
}

@article{consistency,
  title={Consistency policy: Accelerated visuomotor policies via consistency distillation},
  author={Prasad, Aaditya and Lin, Kevin and Wu, Jimmy and Zhou, Linqi and Bohg, Jeannette},
  journal={arXiv preprint arXiv:2405.07503},
  year={2024}
}

@article{var,
  title={Variational distillation of diffusion policies into mixture of experts},
  author={Zhou, Hongyi and Blessing, Denis and Li, Ge and Celik, Onur and Jia, Xiaogang and Neumann, Gerhard and Lioutikov, Rudolf},
  journal={Advances in Neural Information Processing Systems},
  volume={37},
  pages={12739--12766},
  year={2024}
}

@article{mean,
  title={Mean flows for one-step generative modeling},
  author={Geng, Zhengyang and Deng, Mingyang and Bai, Xingjian and Kolter, J Zico and He, Kaiming},
  journal={arXiv preprint arXiv:2505.13447},
  year={2025}
}

@inproceedings{habitat,
  title     =     {Habitat: {A} {P}latform for {E}mbodied {AI} {R}esearch},
  author    =     {Manolis Savva and Abhishek Kadian and Oleksandr Maksymets and Yili Zhao and Erik Wijmans and Bhavana Jain and Julian Straub and Jia Liu and Vladlen Koltun and Jitendra Malik and Devi Parikh and Dhruv Batra},
  booktitle =     {Proceedings of the IEEE/CVF International Conference on Computer Vision (ICCV)},
  year      =     {2019}
}

@article{ddpm,
  title={Denoising diffusion probabilistic models},
  author={Ho, Jonathan and Jain, Ajay and Abbeel, Pieter},
  journal={Advances in neural information processing systems},
  volume={33},
  pages={6840--6851},
  year={2020}
}

@article{navidiffusor,
  title={NaviDiffusor: Cost-Guided Diffusion Model for Visual Navigation},
  author={Zeng, Yiming and Ren, Hao and Wang, Shuhang and Huang, Junlong and Cheng, Hui},
  journal={arXiv preprint arXiv:2504.10003},
  year={2025}
}

@inproceedings{prior,
  title={Prior does matter: Visual navigation via denoising diffusion bridge models},
  author={Ren, Hao and Zeng, Yiming and Bi, Zetong and Wan, Zhaoliang and Huang, Junlong and Cheng, Hui},
  booktitle={Proceedings of the Computer Vision and Pattern Recognition Conference},
  pages={12100--12110},
  year={2025}
}

@article{ddbm,
  title={Denoising diffusion bridge models},
  author={Zhou, Linqi and Lou, Aaron and Khanna, Samar and Ermon, Stefano},
  journal={arXiv preprint arXiv:2309.16948},
  year={2023}
}

@article{flow,
  title={Flow matching for generative modeling},
  author={Lipman, Yaron and Chen, Ricky TQ and Ben-Hamu, Heli and Nickel, Maximilian and Le, Matt},
  journal={arXiv preprint arXiv:2210.02747},
  year={2022}
}

@article{knowledge,
  title={Knowledge diffusion for distillation},
  author={Huang, Tao and Zhang, Yuan and Zheng, Mingkai and You, Shan and Wang, Fei and Qian, Chen and Xu, Chang},
  journal={Advances in Neural Information Processing Systems},
  volume={36},
  pages={65299--65316},
  year={2023}
}

@article{onem,
  title={One-step diffusion distillation through score implicit matching},
  author={Luo, Weijian and Huang, Zemin and Geng, Zhengyang and Kolter, J Zico and Qi, Guo-jun},
  journal={Advances in Neural Information Processing Systems},
  volume={37},
  pages={115377--115408},
  year={2024}
}

@article{progressive,
  title={Progressive distillation for fast sampling of diffusion models},
  author={Salimans, Tim and Ho, Jonathan},
  journal={arXiv preprint arXiv:2202.00512},
  year={2022}
}

@article{consistency1,
  title={Consistency models},
  author={Song, Yang and Dhariwal, Prafulla and Chen, Mark and Sutskever, Ilya},
  year={2023}
}

@article{ppo,
  title={Proximal policy optimization algorithms},
  author={Schulman, John and Wolski, Filip and Dhariwal, Prafulla and Radford, Alec and Klimov, Oleg},
  journal={arXiv preprint arXiv:1707.06347},
  year={2017}
}

@inproceedings{sac,
  title={Soft actor-critic: Off-policy maximum entropy deep reinforcement learning with a stochastic actor},
  author={Haarnoja, Tuomas and Zhou, Aurick and Abbeel, Pieter and Levine, Sergey},
  booktitle={International conference on machine learning},
  pages={1861--1870},
  year={2018},
  organization={Pmlr}
}

@article{dppo,
  title={Diffusion policy policy optimization},
  author={Ren, Allen Z and Lidard, Justin and Ankile, Lars L and Simeonov, Anthony and Agrawal, Pulkit and Majumdar, Anirudha and Burchfiel, Benjamin and Dai, Hongkai and Simchowitz, Max},
  journal={arXiv preprint arXiv:2409.00588},
  year={2024}
}

@inproceedings{fdpp,
  title={Fdpp: Fine-tune diffusion policy with human preference},
  author={Chen, Yuxin and Jha, Devesh K and Tomizuka, Masayoshi and Romeres, Diego},
  booktitle={2025 IEEE International Conference on Robotics and Automation (ICRA)},
  pages={12010--12016},
  year={2025},
  organization={IEEE}
}

@article{ncdpo,
  title={Fine-tuning Diffusion Policies with Backpropagation Through Diffusion Timesteps},
  author={Yang, Ningyuan and Gao, Jiaxuan and Gao, Feng and Wu, Yi and Yu, Chao},
  journal={arXiv preprint arXiv:2505.10482},
  year={2025}
}

@article{reinflow,
  title={ReinFlow: Fine-tuning flow matching policy with online reinforcement learning},
  author={Zhang, Tonghe and Yu, Chao and Su, Sichang and Wang, Yu},
  journal={arXiv preprint arXiv:2505.22094},
  year={2025}
}

@inproceedings{efficentnet,
  title={Rethinking model scaling for convolutional neural networks},
  author={Tan, Mingxing and Le, Q Efficientnet and others},
  booktitle={Proceedings of the International conference on machine learning, Long Beach, CA, USA},
  volume={15},
  year={2019}
}

@article{attention,
  title={Attention is all you need},
  author={Vaswani, Ashish and Shazeer, Noam and Parmar, Niki and Uszkoreit, Jakob and Jones, Llion and Gomez, Aidan N and Kaiser, {\L}ukasz and Polosukhin, Illia},
  journal={Advances in neural information processing systems},
  volume={30},
  year={2017}
}

@inproceedings{film,
  title={Film: Visual reasoning with a general conditioning layer},
  author={Perez, Ethan and Strub, Florian and De Vries, Harm and Dumoulin, Vincent and Courville, Aaron},
  booktitle={Proceedings of the AAAI conference on artificial intelligence},
  volume={32},
  number={1},
  year={2018}
}

@inproceedings{unet,
  title={U-net: Convolutional networks for biomedical image segmentation},
  author={Ronneberger, Olaf and Fischer, Philipp and Brox, Thomas},
  booktitle={International Conference on Medical image computing and computer-assisted intervention},
  pages={234--241},
  year={2015},
  organization={Springer}
}

@article{gostanford,
  title={Deep visual mpc-policy learning for navigation},
  author={Hirose, Noriaki and Xia, Fei and Mart{\'\i}n-Mart{\'\i}n, Roberto and Sadeghian, Amir and Savarese, Silvio},
  journal={IEEE Robotics and Automation Letters},
  volume={4},
  number={4},
  pages={3184--3191},
  year={2019},
  publisher={IEEE}
}

@article{depth_any_thing,
  title={Depth anything v2},
  author={Yang, Lihe and Kang, Bingyi and Huang, Zilong and Zhao, Zhen and Xu, Xiaogang and Feng, Jiashi and Zhao, Hengshuang},
  journal={Advances in Neural Information Processing Systems},
  volume={37},
  pages={21875--21911},
  year={2024}
}

@article{navdp,
  title={Navdp: Learning sim-to-real navigation diffusion policy with privileged information guidance},
  author={Cai, Wenzhe and Peng, Jiaqi and Yang, Yuqiang and Zhang, Yujian and Wei, Meng and Wang, Hanqing and Chen, Yilun and Wang, Tai and Pang, Jiangmiao},
  journal={arXiv preprint arXiv:2505.08712},
  year={2025}
}

@inproceedings{gibson,
  title     = {Gibson Env: Real-World Perception for Embodied Agents},
  author    = {Xia, Fei and Zamir, Amir R. and He, Zhiyang and Sax, Alexander and Malik, Jitendra and Savarese, Silvio},
  booktitle = {Proceedings of the IEEE/CVF Conference on Computer Vision and Pattern Recognition (CVPR)},
  pages     = {9068--9079},
  year      = {2018}
}

@article{mp3d,
  title={Matterport3d: Learning from rgb-d data in indoor environments},
  author={Chang, Angel and Dai, Angela and Funkhouser, Thomas and Halber, Maciej and Niessner, Matthias and Savva, Manolis and Song, Shuran and Zeng, Andy and Zhang, Yinda},
  journal={arXiv preprint arXiv:1709.06158},
  year={2017}
}

@article{scand,
  title={Socially compliant navigation dataset (scand): A large-scale dataset of demonstrations for social navigation},
  author={Karnan, Haresh and Nair, Anirudh and Xiao, Xuesu and Warnell, Garrett and Pirk, S{\"o}ren and Toshev, Alexander and Hart, Justin and Biswas, Joydeep and Stone, Peter},
  journal={IEEE Robotics and Automation Letters},
  volume={7},
  number={4},
  pages={11807--11814},
  year={2022},
  publisher={IEEE}
}

@article{distfm,
  title={Score Distillation of Flow Matching Models},
  author={Zhou, Mingyuan and Gu, Yi and Zheng, Huangjie and Song, Liangchen and He, Guande and Zhang, Yizhe and Hu, Wenze and Yang, Yinfei},
  journal={arXiv preprint arXiv:2509.25127},
  year={2025}
}

\end{document}